\documentclass[final,3p,times,authoryear]{elsarticle}
\usepackage{setspace}
\usepackage{threeparttable}
\usepackage[draft]{changes}
\usepackage[inline]{enumitem}
\definechangesauthor[color=blue]{PC}
\usepackage{subcaption}
\usepackage{lineno,hyperref}
\usepackage{amsmath,amssymb,amsfonts}
\usepackage{algorithmic}
\usepackage{graphicx}
\usepackage{anyfontsize}
\usepackage{adjustbox}
\usepackage{rotating}
\usepackage{tablefootnote}
\usepackage{threeparttable}
\usepackage{longtable}
\usepackage{tabularray}
\usepackage{caption}
\usepackage[font=small,labelfont=bf, figurename=Fig.]{caption} 

\usepackage{textcomp}
\usepackage{todonotes}
\usepackage{tabularx}
\usepackage{gensymb}
\usepackage[utf8]{inputenc}
\usepackage{mathtools, nccmath}
\usepackage{booktabs}
\usepackage{cleveref}
\usepackage{chemformula}
\usepackage[version=4]{mhchem}
\usepackage{booktabs, multirow} 
\usepackage{soul}
\modulolinenumbers[1]

\usepackage{comment}
\usepackage[inline]{enumitem}   
\makeatletter
\newcommand{\inlineitem}[1][]{%
\ifnum\enit@type=\tw@
    {\descriptionlabel{#1}}
  \hspace{\labelsep}%
\else
  \ifnum\enit@type=\z@
       \refstepcounter{\@listctr}\fi
    \quad\@itemlabel
    \hspace{\labelsep}%
\fi}
\makeatother
\begin{document}

\begin{frontmatter}

\title{SGAP-Gaze: Scene-Grid Attention Based Point-of-Gaze Estimation Framework for Driver Gaze}





\author[mymainaddress]{Pavan Kumar Sharma}


\author[mymainaddress]{Pranamesh Chakraborty\corref{mycorrespondingauthor}}



\cortext[mycorrespondingauthor]{Corresponding author\\
Pavan Kumar Sharma: pavans20@iitk.ac.in, Pranamesh Chakraborty: pranames@iitk.ac.in, +91-512-259-2146}

\address[mymainaddress]{Department of Civil Engineering, Indian Institute of Technology Kanpur, Kanpur-208016, U.P., India}



\begin{abstract}
Driver gaze estimation is essential for understanding drivers' visual attention and situational awareness of the surrounding traffic. Existing gaze estimation models primarily rely on driver facial information to predict the Point-of-Gaze (PoG) or the 3D gaze direction vector, often without explicitly incorporating the visual context of the driving scene. The scene images provide cues about surrounding traffic, which can help improve the gaze estimation model, along with the face images. 
We propose \textbf{SGAP-Gaze}, \textbf{S}cene-\textbf{G}rid \textbf{A}ttention based \textbf{P}oint-of-\textbf{Gaze} estimation network, trained and tested on our UD-FSG dataset, which explicitly incorporates the scene images into the gaze estimation modeling. The proposed network integrates driver face, eye, iris, and scene contextual information. First, the features extracted from facial modalities are fused to form a gaze-intent vector. Then, attention scores are computed over the spatial scene grid using a Transformer-based attention mechanism fusing face and scene image features to obtain the PoG. The proposed SGAP-Gaze model achieved mean pixel errors of 96.16 on the UD-FSG dataset and 63.48 on the Look Both Ways (LBW) dataset, corresponding to 6.55\% and 5.98\% of the respective scene-image diagonals. On the UD-FSG dataset, SGAP-Gaze reduces the mean pixel error by 29.78\% compared to state-of-the-art driver gaze estimation models. The spatial pixel distribution analysis shows that SGAP-Gaze consistently achieves lower mean pixel error than existing methods across all spatial ranges, including the outer regions of the scene, which are rare but critical for understanding driver attention. These results highlight the effectiveness of integrating multi-modal gaze cues with scene-aware attention for a robust driver PoG estimation model in real-world driving environments.
\end{abstract}

\end{frontmatter}



\section{Introduction}
Road user safety remains a major global concern, particularly in developing countries like India, where heterogeneous traffic conditions and poor lane discipline significantly increase safety risks. In such environments, smaller vehicles (e.g., motorbikes and autorickshaws) typically do not strictly follow lane discipline and frequently maneuver through gaps to overtake other vehicles. This unpredictable traffic behavior makes driving more challenging for all other road users \citep{liu2025beyond}. Consequently, safe driving in these conditions heavily depends on the driver’s ability to continuously monitor the surrounding environment and rapidly allocate visual attention to relevant regions of the scene \citep{wang2026driver}. Inattentive driving in such a complex traffic environment may increase the likelihood of crashes or near misses \citep{regan2011driver, yuan2022self}.
According to the Ministry of Road Transport and Highways (MoRTH, 2023), nearly 4.8 lakh road accidents occurred in India in 2023 \citep{morth2023roadaccidents}. Although several factors contribute to these accidents, one of them is the driver's lack of situational awareness of the surrounding traffic \citep{dong2010driver, li2019drivers, li2023much}. A driver monitoring system that monitors the driver's visual attention using gaze information (gaze position in the scene, fixation time, duration, etc.) \citep{wu2019gaze, mole2021drivers} can help to improve driver safety by alerting the driver. Although the automobile industry has made significant advancements in driver safety features, including Driver Monitoring Systems (DMS), these systems are primarily designed to detect in-cabin distractions and typical functionalities include identifying behaviors such as mobile phone usage (talking or texting), head pose deviation, and drowsiness detection (based on blink rate or eye closure patterns) \citep{dong2010driver, koesdwiady2016recent, lin2022innovative, wang2022driver, michelaraki2023real}. While such systems improve driver safety, they primarily focus on external driver states rather than assessing whether the driver is attending to the surrounding traffic environment. An efficient driver gaze-monitoring system has the potential to significantly improve driver safety.

\par A driver gaze monitoring system first requires estimating driver gaze accurately, which is then used to measure visual attention. Driver gaze estimation is typically performed by mounting one or more cameras on the dashboard to determine where the driver is gazing. This non-intrusive gaze-monitoring enables scalable deployment in the real world, unlike the intrusive method, which requires the driver to wear an eye tracker \citep{sharma2024review, sharma2025evaluation}. Driver gaze representation is also an important aspect of gaze estimation. In the literature, driver gaze is typically represented in three common ways: (a) zone-based gaze classification, (b) gaze vector and/or Point of Gaze (PoG), and (c) object-based gaze estimation \citep{sharma2024review}. In the zone-based approach, the windshield of a car is divided into predefined zones, such as the forward, rearview mirror, and left/right side mirrors \citep{fridman2016driver, sharma2024driver}. The gaze estimation model predicts the driver's gaze using driver face features and classifies it into predefined zones, such as forward and rearview mirror. This approach of gaze representation is widely used in the literature because of its simplicity. However, it provides only a coarse representation of gaze. In the gaze vector–based approach, gaze is represented as a 3D gaze vector originating from the driver's eyes. Model performance is evaluated by computing the angular error between the predicted gaze vector and the ground truth gaze vector \citep{hosp2020remoteeye, jha2021multimodal,  kasahara2022look, cheng2024you, hu2025lnet, zhang2026glgaze}. When this vector is projected onto the driving scene, it indicates the location (point) on the scene at which the driver is looking, and is also known as the Point of Gaze (PoG) \citep{yuan2022self}. In some studies, a visual scene saliency map is also generated on the scene image to indicate the approximate gaze location within the scene in the form of a heatmap \citep{alletto2016dr, kasahara2022look}.
The third approach of gaze representation is the object-based gaze estimation, which aims to predict the gaze object in the scene that the driver is looking at \citep{dua2020dgaze}. Unlike the zone-based approach, this method explicitly incorporates scene information and identifies the relevant object of attention. Please note that the PoG can be used to determine the object on which the driver is gazing by using an additional object detection model on the driving scene. Therefore, the PoG method can be considered the most comprehensive approach for representing driver gaze. In this study, we have primarily focused on PoG estimation. To compare our proposed model with state-of-the-art methods, 3D gaze direction estimation has also been used.

\par The 3D gaze vector or PoG is typically estimated using cameras facing the driver. The existing open-source driver gaze datasets rely on driver-facing cameras and only a few provide ground-truth gaze vectors. The majority of existing datasets are zone-based and collected in a parked-vehicle environment. Although driver face is the primary requirement for estimating driver gaze, the gaze estimation can be made more robust and accurate if scene information is present \citep{li2019drivers, lemonnier2020drivers, kaya2021hey}. To the best of the authors' knowledge, only one dataset, LBW (Look Both Ways), includes both face and scene images,  collected in homogeneous real-world driving scenarios \citep{kasahara2022look}. However, the scene in the dataset is limited to very few vehicles or traffic agents in the surrounding traffic. Therefore, there is a need for a large-scale driver gaze dataset with driver face and scene camera data collected in a dynamic driving environment with enough traffic agents present. This can help to build a state-of-the-art, robust driver gaze estimation model. 
\par Since existing driver gaze datasets are based on cameras facing the driver, the gaze estimation models are developed based on the driver face information only \citep{vora2018driver, ghosh2021speak2label, wu2025multi}. Even though the LBW dataset contains both face and scene information, the scene information is used to create a scene saliency map rather than being used directly as input for gaze estimation \citep{kasahara2022look}. However, studies on driver visual behavior have shown that gaze is influenced by the surrounding scene, such as the positions of vehicles \citep{li2019drivers, lemonnier2020drivers, kaya2021hey}. Therefore, incorporating scene information can improve gaze estimation models. To address these limitations, we have collected a diverse real-world driving gaze dataset called \textbf{U}rban \textbf{D}riving\textbf{-F}ace \textbf{S}cene \textbf{G}aze (\textbf{UD-FSG}), in heterogeneous traffic environments. It comprises synchronous driver face images, scene images, and eye-tracker data to obtain gaze ground truth on the scene image. Using this dataset, we propose a gaze estimation model that directly incorporates scene information and driver facial information as inputs. The driver's face, eye, and iris are first detected using a custom YOLOv8 (You Only Look Once version 8) Face-Eye-Iris (FEI) detector trained on the data of 481 different drivers' face images. Then, we extract face features and iris-weighted eye features using a pretrained ResNet (Residual Networks) convolutional neural network (CNN) model. On the other hand, the driving scene is processed using a ResNet-based feature extractor to obtain spatially distributed feature representations, where each feature corresponds to a localized region (grid) of the scene. The extracted features from facial modalities are then fused to form a gaze intent vector. Attention scores are computed over spatial scene grids using a Transformer-based attention mechanism that fuses both face and scene information. The point of gaze (PoG) is then obtained as the expectation of the scene grid centers under the normalized attention distribution. Our proposed methodology is then compared with state-of-the-art gaze estimation models using our

own UD-FSG dataset. We have also tested our model on the benchmark LBW dataset to demonstrate its performance. Overall, the key contributions of our study are as follows:
\begin{enumerate}[itemsep=0pt,parsep=0pt,topsep=0pt,partopsep=0pt]
\item An improved eye feature representation using an iris-weighted Gaussian function, which emphasizes the iris region and helps capture the iris movement pattern within the eye.
\item We propose a scene grid attention-based point of gaze (SGAP-Gaze) estimation network that explicitly incorporates the scene information along with the driver's face for gaze estimation purposes.
\item Overall gaze estimation metrics provide limited insight into the spatial variation of prediction errors, particularly in peripheral and less frequently attended regions of the driving scene. This motivates a more detailed evaluation of gaze estimation performance across different spatial ranges.
\end{enumerate}

\par The rest of the paper is organized as follows. Section 2 provides a relevant literature overview on gaze estimation and driver attention prediction. Section 3 describes the datasets used in this study, followed by Section 4, which presents the proposed methodology, including the complete pipeline of the gaze estimation model and its implementation details. Section 5 reports the experimental results, including a comparison of model performance with state-of-the-art methods. Finally, we conclude with the key findings and discuss the strengths and limitations of the proposed gaze-estimation approach


\section{Related Work}
Driver gaze estimation is the first step towards developing a model for driver visual attention monitoring. Continuous monitoring of driver visual attention while driving helps the drivers to improve their safety. The need for continuous monitoring of driver visual attention further increases, particularly during complex traffic maneuvers such as turning left or right, going straight through intersections, and entering or exiting on-ramps. Although some leading automobile manufacturers have introduced advanced driving assistance systems (ADAS) in their cars for driver assistance and safety, specifically for driver monitoring, these systems are primarily limited to detecting driver distraction and drowsiness \citep{dong2010driver, koesdwiady2016recent, lin2022innovative, michelaraki2023real}. There is significant scope for research on monitoring driver visual attention, specifically the influence of surrounding traffic conditions, and during the performance of different complex driving maneuvers.
\par Intrusive and non-intrusive methods are two broad categories of gaze monitoring in the literature based on the placement and use of cameras \citep{sharma2024review}. Intrusive methods typically involve head-mounted devices or wearable eye trackers that directly record eye movements and gaze information \citep{dukic2012older, romoser2009effect, li2019drivers, lemonnier2020drivers}. These methods provide high accuracy but are often impractical for large-scale real-world deployment due to user discomfort and setup constraints. In contrast, non-intrusive methods rely on remotely mounted cameras installed inside the vehicle, such as on the dashboard, windshield, or interior roof \citep{chuang2014estimating, vicente2015driver, yoon2019driver, sharma2025evaluation}. These systems use single or multiple cameras along with machine learning or deep learning models/algorithms for driver gaze estimation. These methods are more practical and scalable than intrusive ones. Therefore, non-intrusive methods are commonly adopted in modern driver monitoring systems. Apart from the estimation methods of driver gaze, the representation of gaze is also an important aspect in driver gaze estimation.
In the literature, it is represented in three common ways: (a) Gaze zone/gaze class/area of interest, (b) Gaze direction vector or point of gaze (PoG), and (c) Gaze object \citep{sharma2024review}. The most commonly used representation of driver gaze is in terms of gaze zone. Here, gaze zones are predefined regions on the windshield or dashboard of the cars, typically including the forward roadway, rearview mirror, speedometer/instrument cluster, left and right-side mirrors, dashboard, and side windows \citep{vicente2015driver, vasli2016driver, tawari2014attention, fridman2016driver}. The gaze estimation model classifies the driver's gaze in one of the gaze zones in which the driver is looking. In recent years, some studies reported driver gaze in a finer way in terms of gaze direction vector \citep{kasahara2022look, hu2025lnet, zhang2026glgaze} and/or point of gaze (PoG) \citep{yang2019dual}. In gaze estimation, gaze vector-based model performance is measured using the mean angular error in degrees, whereas in PoG, it is measured using pixel error with respect to the scene image. The third approach represents gaze in terms of a gaze object. Here, gaze object is the object at which driver is looking while driving such as a car, bus, pedestrians, traffic lights, road signs and the background \citep{dua2020dgaze}. This approach of gaze representation is underexplored and has potential research scope. Only a limited number of studies—primarily in simulated driving environments \cite{dua2020dgaze} have attempted to represent driver gaze in terms of gaze objects in the scene. 
\par The gaze vector and/or point of gaze (PoG) is commonly modeled using probabilistic \citep{kasahara2022look}, deep learning, and more recently, transformer-based approaches \citep{jha2023driver, hu2025lnet}. Several studies estimate driver gaze by regressing facial and/or eye features, while few studies incorporate scene information \cite{hu2025lnet} as an additional spatial input alongside face and eye features. \citet{yang2019dual} developed a dual-camera-based driver gaze estimation model in which a front dashboard camera captures the driver's face, and a camera mounted on the vehicle's interior roof captures the vehicle interior and windshield. They used a Volterra Non-linear Regressive with Exogenous inputs (VNRX) model, where the inputs are the face features and the outputs are the point of gaze on the camera 2 image.
Similarly, \citet{yuan2022self} proposed a self-calibrated gaze estimation system that eliminates manual calibration. It combines feature extraction from facial landmarks, head pose, and eye movements with unscented Kalman filtering for reliable tracking. Gaze pattern learning is then mapped to different zones using Gaussian Process Regression. For gaze estimation in terms of gaze vector, \citet{cheng2024you} proposed GazePTR (Gaze Pyramid Transformer) to address low-resolution face images, which takes a full-face image as input, from which a CNN backbone extracts hierarchical feature maps at multiple levels. Instead of relying only on the final feature map, the model collects multi-level face features, aligns them using 1×1 convolutions and global average pooling to maintain consistent dimensions, and feeds them into a transformer encoder. A learnable token aggregates cross-level information for gaze estimation. To ensure that low-level features (particularly eye-region details) are effectively utilized, the authors introduce auxiliary gaze supervision at each pyramid level, improving robustness under extreme head poses and low-resolution conditions. This method relies only on infrared face images and does not incorporate separate scene images, limiting its ability to perform scene-aware visual attention.

\par Some recent studies estimate the gaze vector (gaze direction) and represent the driver's visual attention or focus in the scene by creating a scene saliency map \citep{hu2021data, kasahara2022look, jha2023driver, xu2025modeled}. The study by \citet{hu2021data} used the SalGAN adversarial framework for saliency map estimation, incorporating an element-wise sigmoid to interpret each pixel as a probability. Their architecture combined bottleneck, multi-resolution, and transition modules, enabling both down-sampling and up-sampling to enrich high-resolution features with global context. In contrast, this design strengthens feature fusion for dual-view gaze and scene representation. This study's reliance on a driving simulator, limits its applicability in real-world driving conditions. A study by \citet{kasahara2022look} developed a self-supervised algorithm to estimate the driver gaze. The model takes a driver's face image and a scene as input and outputs 3D gaze direction and scene saliency. This study used the ETH-XGaze model based on ResNet-50 for gaze vector and the Unisal (MNetV2-RNN-Decoder) for scene saliency map. The study by \citet{jha2023driver} uses head pose features (3D position and orientation: yaw, pitch, roll) along with eye appearance features extracted from cropped eye images. They propose a two-branch deep learning model in which a fully connected head-pose encoder and a CNN-based eye encoder are fused, followed by a multi-stage upsampling decoder. Instead of directly regressing coordinates, the model predicts a probabilistic 2D visual attention heatmap over the scene, where the ground truth gaze is represented as a spatial probability distribution. 
\par Overall, most existing studies rely on single-level facial features for gaze estimation, whereas few studies explore multi-level facial-feature fusion \citep{cheng2024you}, which can better capture head movement orientation. Additionally, many approaches overlook the role of iris position, an important cue, particularly when the driver's gaze is away from the forward regions of the scene. Furthermore, existing gaze estimation methods predominantly rely on facial features and do not explicitly incorporate scene context except a study by \cite{hu2025lnet}. A few studies use scene information to compute scene saliency via a heatmap rather than directly using it as an input to a gaze estimation model. To address these limitations, we propose a model that integrates multi-level facial features and iris information, and explicitly incorporates scene context to enable precise driver gaze estimation.

\section{Dataset}
Deep-learning-based gaze estimation models, particularly those relying on convolutional neural networks (CNNs) and vision transformers, require large-scale, diverse driver-face data, or face data together with scene data to achieve robust performance in driver gaze estimation tasks. Data collection in real driving environments presents several practical challenges, including reflections from sunglasses, glare due to sunlight, and poor illumination under low-light conditions. Such factors are typically absent or inadequately represented in simulated driving environments. In this study, we separately train and test our model on two different datasets to evaluate the robustness and generalization capability of the proposed gaze estimation model. The first dataset is our UD-FSG (Urban Driving-Face Scene Gaze) dataset \cite{} and the second is the existing benchmark LBW (Look Both Ways) dataset \citep{kasahara2022look}. In this section, we provide a detailed description of our UD-FSG dataset including the sensor setup and synchronization employed in our laboratory instrumented vehicle, characteristics of the dataset, and comparison with the existing driver gaze datasets. We also briefly discuss the key details of the existing benchmark LBW dataset. 

\subsection{UD-FSG Dataset}
\subsubsection{Sensor Setup and Synchronization}
The data were collected using Transportation Engineering Laboratory Instrumented Vehicle. Two web cameras were mounted inside the instrumented vehicle on the dashboard, one in front of the driver's seat to capture the driver face image and the other at the center of the dashboard to capture driver's forward traffic scene-image. Both cameras were synchronized using a C++ GStreamer application. An eye tracker was used to obtain the gaze ground truth and was synchronized with the cameras using timestamps based on stopwatch time. 

\subsubsection{Participants}
The present study requires driver face and gaze information in addition to the forward traffic scene to build the driver gaze estimation model. Since this study involves human participants (drivers), ethical approval was obtained from the \textit{Institute Ethics Committee} before conducting the study. Each participant provided informed consent by signing a consent form before the data collection process began. This study recruited 41 male professional drivers to drive the instrumented vehicle. However, due to issues with the data storage device, data from 6 participants were not included, resulting in a final dataset comprising 35 participants. The dataset includes the following driver statistics: mean age = 35.77 years (SD = 6.30) and mean driving experience = 13.7 years (SD = 5.85).

\subsubsection{Dataset Details}
The data were collected in Kanpur city, India, in a heterogeneous traffic environment across arterial and sub-arterial roads to incorporate variations in traffic density in the dataset. The driver's face and the forward traffic scene were recorded using two cameras; one mounted in front of the driver's seat and another mounted at the center of the dashboard to capture the respective images simultaneously. To obtain the gaze ground truth, an eye tracker was also used, and the gaze coordinates were subsequently transformed into the coordinate system of the scene camera images. Further details of gaze ground truth transformation from the eye-tracker scene-camera frame to the forward-facing scene camera are provided in our another study \citep{sharma2026transgaze}. Data collection was conducted at different times of day to account for the effects of varying sunlight conditions on the driver's face. Each driver drove approximately one hour, covering a travel distance of about 30–35 km. The data were originally recorded at 10 frames per second (FPS), and frames were extracted at 5 FPS to create this dataset. Therefore, the Urban Driving–Face Scene Gaze (UD-FSG) dataset is a large-scale benchmark real driving gaze dataset comprising data from 35 drivers, which includes 373,488 driver face and scene image pairs, along with scene traffic-object bounding-box coordinates and gaze labels in terms of 2D gaze ground truth coordinates with respect to the scene image and the gaze object. The gaze-object label identifies the object at which the driver is looking. 
\par
\begin{figure}[!htb]
    \centering
    \includegraphics[width=1.0\textwidth]{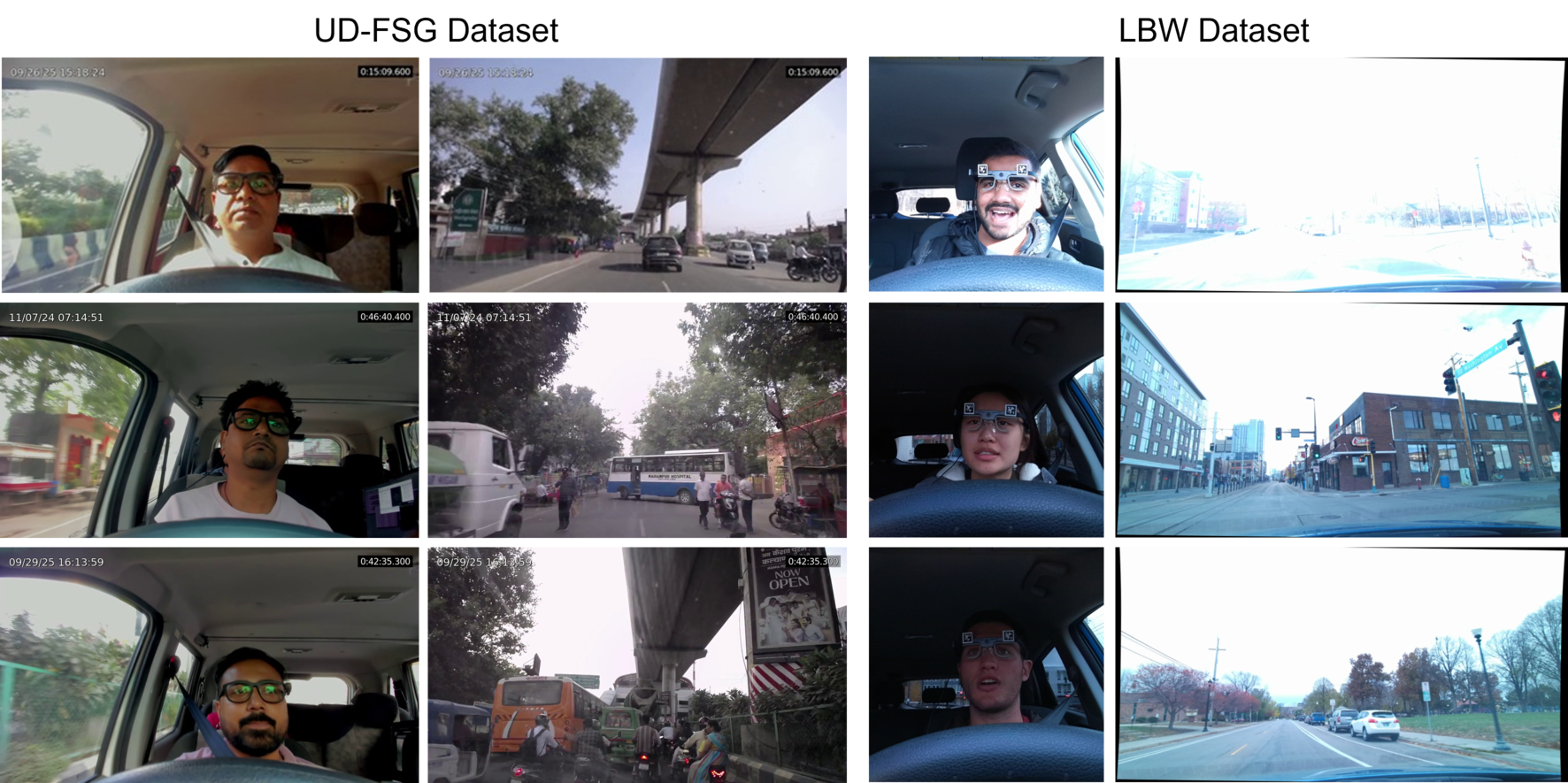}
        \caption{Sample of face–scene image pairs from the UD-FSG and LBW datasets.}
        \label{fig:data_sample}
\end{figure}

 Our dataset is significantly larger than the LBW dataset and includes more variation in traffic density and lighting conditions, as shown in Figure \ref{fig:data_sample}. The traffic environment consisting of diverse dynamic traffic agents (vehicles, pedestrians, etc.) makes the scene information meaningful and challenging enough to develop a robust driver gaze estimation model. Also, the eye tracker looks like regular prescription glasses, making the face image similar to that observed during real-world driving tasks. These data have been used to train our gaze estimation model as discussed in the Methodology section. 
\par
\begin{figure}[!htb]
   \centering
        \includegraphics[width=1.0\textwidth]{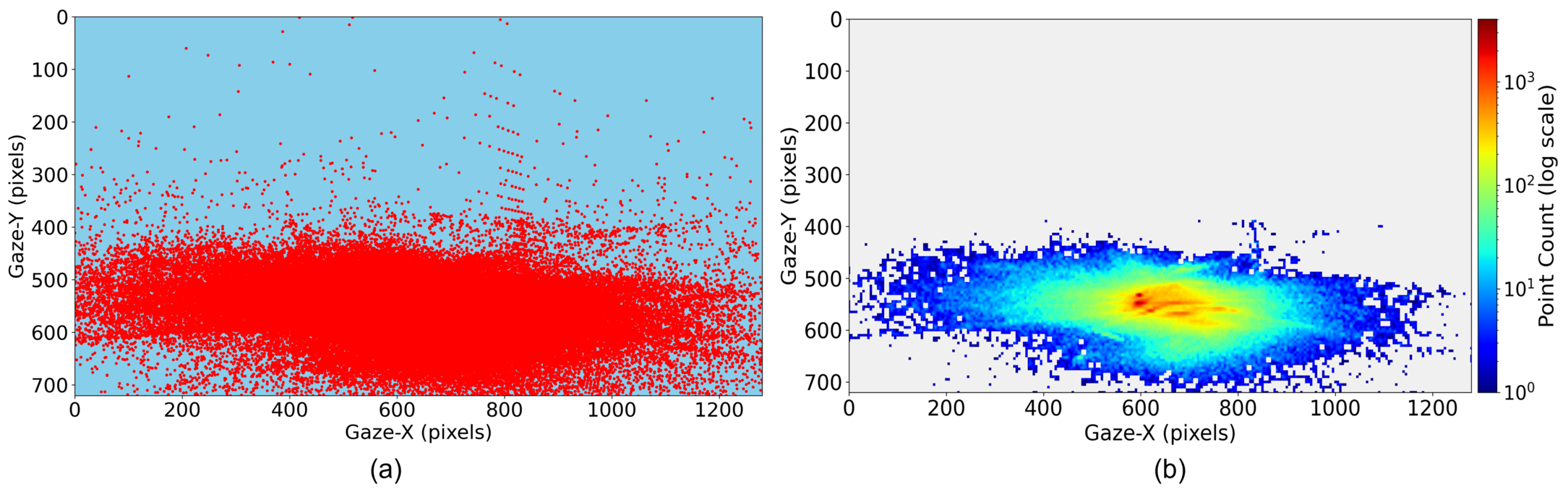}
        \caption{Visualization of driver gaze point distribution of UD-FSG dataset: (a) Gaze point locations on the scene across different drivers (b) Gaze density heatmap computed by aggregating point-of-gaze (PoG) coordinates over 5×5-pixel grid \citep{sharma2026transgaze}}
        \label{fig:data_char}
\end{figure}

 The distribution of driver gaze points in the scene is plotted in Figure \ref{fig:data_char}a by considering the image size equal to the scene image size  ($1280\times720$). To understand the spread of gaze points across the scene, a gaze density heatmap is also plotted by aggregating gaze points over a $5\times5$-pixel region. Figure \ref{fig:data_char}b shows that driver gaze is concentrated in the middle regions (driver forward scene), which is expected, since the driver is predominantly looking in the forward regions while driving. However, our dataset also comprises images in which gaze is distributed across the lateral edges of the scene camera (i.e., $x < 200$ or $x > 1000$ pixels), making it suitable for modeling driver gaze toward the corners of the windshield. The UD-FSG dataset is available at the following link: 
\url{https://github.com/pavans20/Urban-Driving-Face-Scene-Gaze-Dataset.git}.


\subsection{LBW Dataset} Along with the UD-FSG dataset, we have also used the benchmark LBW (Look Both Ways) dataset for comparison of our proposed model with state-of-the-art models. The LBW dataset \citep{kasahara2022look} is a first benchmark driver gaze dataset that contains synchronized face images, scene images, and corresponding gaze ground truth coordinates. The dataset consists of approximately 6.8 hours of real driving data from 28 drivers (22 male and 6 female), totaling 123,297 synchronized face images, scene images, and corresponding 2D and 3D gaze coordinates. For further details about the dataset, the reader can refer to the original paper  \citep{kasahara2022look}.

\section{Methodology}
\begin{figure}[!htbp]
    \centering
    \includegraphics[width=1.0\textwidth]{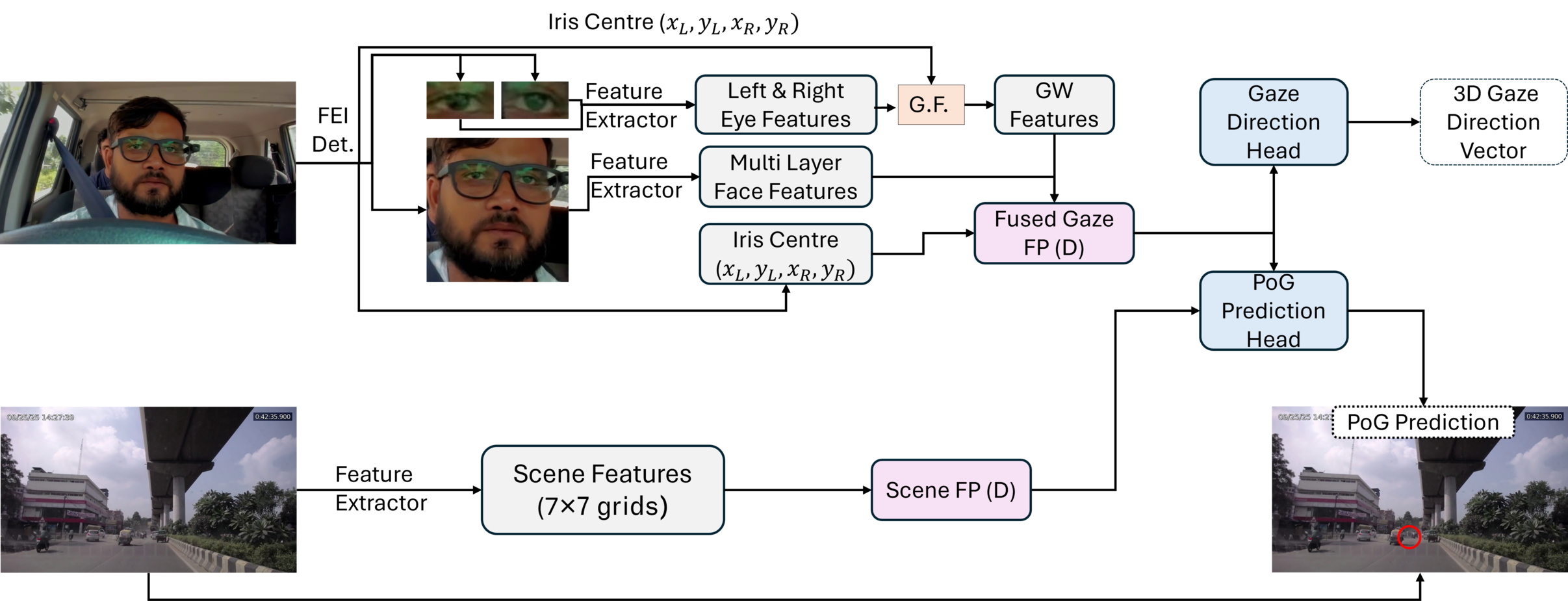}
     \caption{Proposed driver gaze estimation pipeline fusing face and scene information using a transformer.}
    \label{fig:Model Pipeline}
\end{figure}

We propose a scene-grid attention-based framework for driver point-of-gaze (PoG) estimation that jointly leverages facial appearance including face features, eye features, and iris location and scene context. Instead of directly regressing gaze coordinates, the proposed approach formulates gaze estimation as a problem over spatial scene regions. The proposed architecture consists of four major components: (1) Facial geometry detection module responsible for detection of face, eye, iris regions (2) Multi-stream feature extraction module that extracts features from face, eye and scene  (3) Multimodal feature fusion module that integrates heterogeneous feature embeddings into a unified representation, and (4) Gaze prediction head that performs final gaze inference.
Also, to compare with existing 3D gaze directions and state-of-the-art gaze estimation methods, we estimate the 3D gaze direction vector first using only facial features, similar to the features used in the proposed PoG estimation. The pipeline of the proposed driver gaze estimation framework is shown in Figure \ref{fig:Model Pipeline}. 

\textit{Problem Formulation:} Given synchronized inputs of a facial image ($F$) and the corresponding scene image ($S$), the gaze location ($gaze$) is modeled as a function of the face features and scene features:
\begin{equation}
gaze = f(\mathbf{F}, \mathbf{S})
\end{equation}

Since the facial image consists of multiple features at various scales, including eyes, facial feature extraction has been further divided into: (i) Multi-level face features and (ii) Left and right eye features ($\mathbf{e}_L$ and $\mathbf{e}_R$). We next discuss the details of the extraction of facial regions, followed by multi-level feature extraction and feature fusion.

\subsection{Facial Geometry Extraction Module}
The facial image captured from the dashboard face camera contains a face region with context, as shown in Figure \ref{fig:data_sample}. To separate out the context from the face image, we developed a custom Face, Eye, Iris (FEI) detection model to detect the face, eye, and iris regions. A pretrained YOLOv8 model was trained for FEI detection using 2200 annotated face images, obtained from 481 drivers. These data were obtained from different existing zone-based driver gaze datasets \citep{ortega2020dmd, dua2020dgaze, ghosh2021speak2label, sharma2025evaluation}. It is important to note that the data used to train this FEI model are entirely different from our proposed UD-FSG dataset.

\par
\begin{figure}[!htbp]
    \centering
    \includegraphics[width=0.85\textwidth] {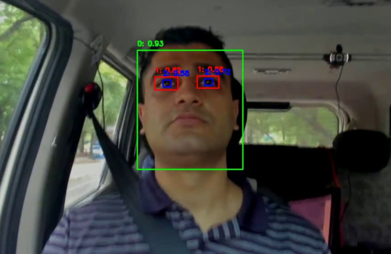}
     \caption{Sample image of detected Face, Eye and Iris using custom YOLOv8 FEI detector}
    \label{fig:/Face_Eye_Iris_Detection}
\end{figure}

 The FEI model achieved mAP (mean average precision) of 95.7\%, at a 0.5 confidence interval and a recall of 93.0\%. Using this FEI model, we first detect the face, eyes, and iris regions from each face image ($F$), as shown in Figure \ref{fig:/Face_Eye_Iris_Detection}. Using the eye and iris bounding box information, we also compute the center coordinates of the iris relative to the top-left corner of the eye for both eyes. In cases where only one iris is detected, we leverage the physiological principle of conjugate eye movement \citep{yarbus2013eye, yang2019dual} to infer the spatial coordinates of the undetected iris from the detected one. Therefore, the final outputs of the facial geometry extraction module include the face region coordinates, the left and right eye region coordinates, and the left and right iris center coordinates relative to their respective eye regions.

\par In some cases, neither iris may be detected because of real-world driving challenges (e.g., occlusion, blur, extreme head pose, or illumination variations). In such cases, we incorporate a validity-based gating mechanism. When reliable iris coordinates are unavailable, the iris input is neutralized, and its contribution is suppressed during feature fusion, preventing noisy or misleading signals from affecting gaze estimation. Consequently, the model adaptively relies on facial appearance, eye features, and scene context, ensuring robust and stable performance even in the absence of iris coordinates. 

\subsection{Multi-Stream Feature Extraction Module}
The multi-stream feature extraction module is designed to process the heterogeneous input modalities through separate feature extraction streams \citep{liu2025beyond}.  Each stream learns modality-specific representations from the face, eye, iris, and scene inputs. 

\subsubsection{Face Feature Extraction}
We employ a pretrained ResNet-18 backbone as a hierarchical feature encoder to obtain compact and discriminative facial representations for gaze estimation. The detected face image is first resized to \(
I_f \in \mathbb{R}^{3 \times 224 \times 224}\) to fit the ResNet architecture input configuration. The resized image is then normalized and passed through the convolutional stem and four residual stages of ResNet-18 to extract intermediate feature maps.
Here, the convolutional stem refers to the initial layers that extract low-level visual features, while the residual stages consist of stacked residual blocks (also called Layer-1/2/3/4) that progressively learn higher-level representations. This can be represented as:

\begin{equation}
F_l = \mathcal{B}_l(I_f), \quad l \in \{1,2,3,4\}
\end{equation}

where, \( \mathcal{B}_l(\cdot) \) denotes the transformation up to the $l^{th}$ residual block,  
\(
F_l \in \mathbb{R}^{C_l \times H_l \times W_l}
\) denotes each feature map
with channel dimensions 
\(
C_l \in \{64,128,256,512\} 
\) for \(
l \in \{1, 2, 3, 4\} 
\), and $H_l$, $W_l$ represent the spatial resolution (height and width) of the feature map.

Since the channel dimensions $C_l$ differ across layers, we project each feature map into a unified 256-dimensional embedding space using a learnable $1 \times 1$ convolution:

\begin{equation}
\hat{F}_l = \phi_l(F_l), \quad 
\hat{F}_l \in \mathbb{R}^{256 \times H_l \times W_l}
\end{equation}

where, $\phi_l(\cdot)$ represents the channel projection operation.

Finally, to obtain a compact global representation, adaptive global average pooling (GAP) is applied over the spatial dimensions to produce a 256-dimensional feature vector from each layer, represented as $f_l = \mathrm{GAP}(\hat{F}_l) \in \mathbb{R}^{256}$.
The response of each channel is computed as:

\begin{equation}
f_l =
\text{GAP}(\hat{F}_l)_c =
\frac{1}{H_l W_l}
\sum_{i=1}^{H_l}
\sum_{j=1}^{W_l}
\hat{F}_l(c,i,j)
\label{eq:gap_feature}
\end{equation}

where, $c \in \{1, \dots, 256\}$ denotes the channel index of the projected feature map.

\begin{figure}[!htbp]
    \centering
    \includegraphics[width=0.85\textwidth] {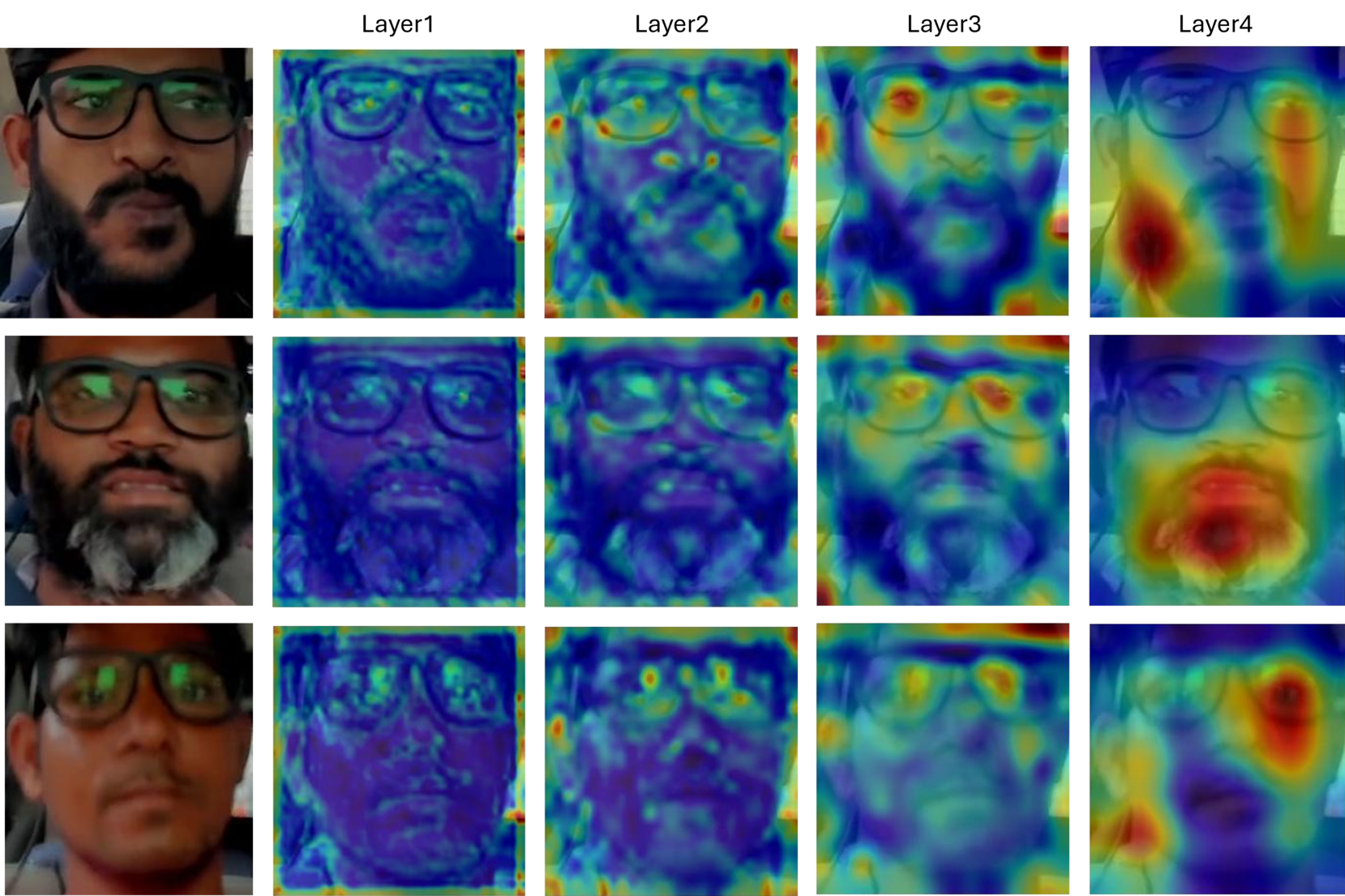}
     \caption{Visualization of face features extracted from different layers of ResNet-18 for three drivers.}
    \label{fig:Different_layer_features}
\end{figure}

Thus, for each face image, four hierarchical global facial feature vectors 
\(
\{f_1, f_2, f_3, f_4\}
\),
each of dimension 256, are extracted from different semantic depths of the network. These multi-level global embeddings capture complementary facial information, ranging from fine-grained texture patterns in shallow layers to high-level structural semantics in deeper layers, as shown in Figure \ref{fig:Different_layer_features}. The resulting 256-dimensional global representations are subsequently utilized in the proposed multi-modal feature fusion module for gaze prediction.

\subsubsection{Gaussian-Weighted Eye Feature Extraction}
\begin{figure}[!htbp]
    \centering
    \includegraphics[width=1.0\textwidth] {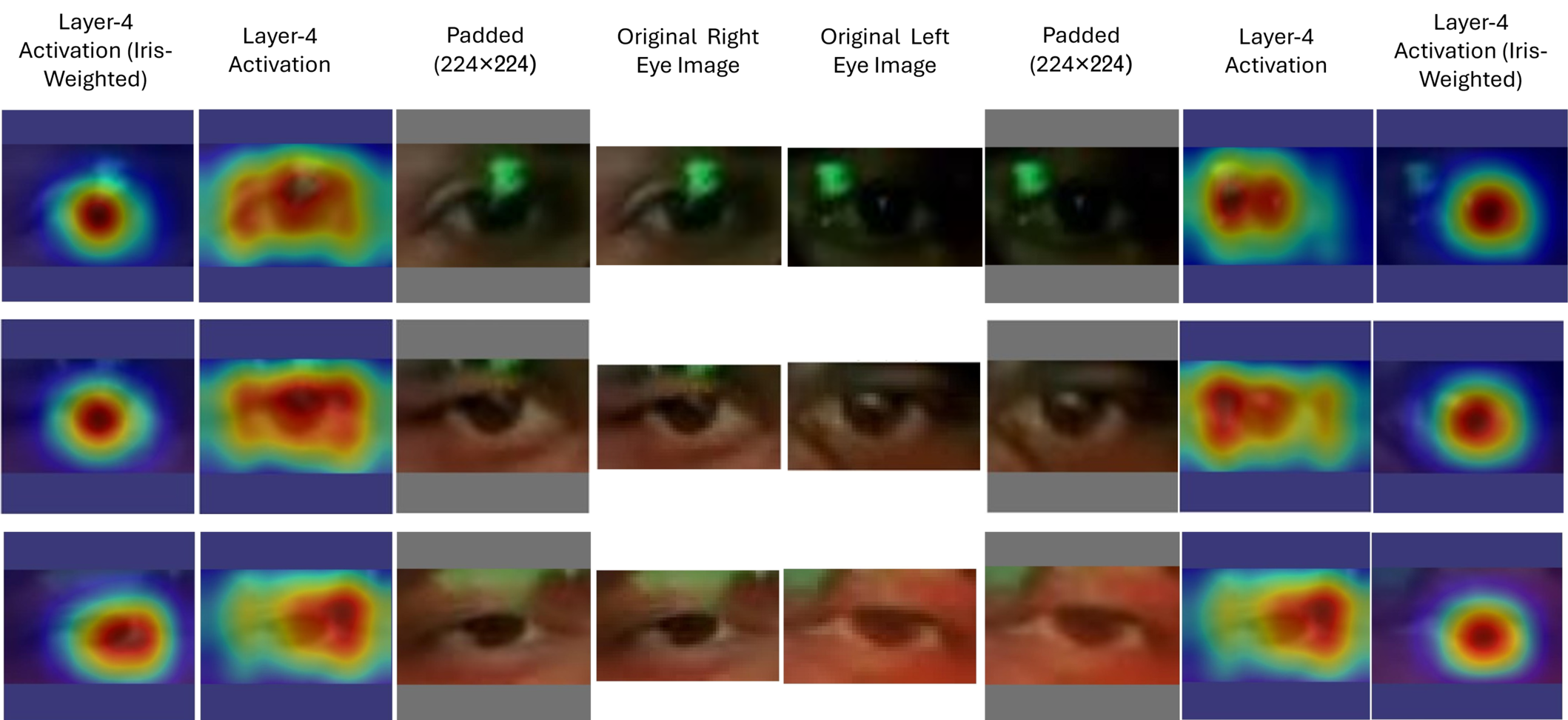}
     \caption{Visualization of eye features extracted from Layer-4 of ResNet-18 with constant padding value of 114, comparing unweighted and Gaussian-weighted feature responses emphasizing the iris region.}
    \label{fig:Eye Features of L4-Layer}
\end{figure}

Along with the overall face features, the eyes, along with the iris position, are extremely important for determining the gaze location of the participants. Therefore, we design an efficient feature extraction of the eye region, detected using our FEI model, along with the iris position. First, the eye features are extracted using a pretrained ResNet-18 backbone, where the final residual block (layer4) is employed to obtain high-level semantic features. However, since the cropped eye images from FEI detection do not meet the required 224×224 input resolution of the ResNet architecture, we applied resizing followed by constant padding to preserve the aspect ratio. A padding value of 114 is selected as a neutral grey intensity to minimise artificial boundary effects and prevent unintended feature activations in the padded regions. Figure \ref{fig:Eye Features of L4-Layer} shows a sample images of the original left and right eye images, along with the padded images used as input to ResNet model. Each padded left and right eye image, with dimensions
\(
I_e \in \mathbb{R}^{3 \times 224 \times 224},
\),
is then normalized and passed through the convolutional stem and residual layers of ResNet-18 to produce a deep feature map for each eye.

\begin{equation}
F_e = \mathcal{B}_4(I_e), \quad 
F_e \in \mathbb{R}^{512 \times H_4 \times W_4}
\end{equation}

Similar to the facial feature extraction module, a $1 \times 1$ convolutional projection is applied to reduce the channel dimensions to 256.

\begin{equation}
\hat{F}_e = \phi_4(F_e), \quad 
\hat{F}_e \in \mathbb{R}^{256 \times H_4 \times W_4}
\end{equation}

To emphasize the iris region derived from the FEI model within the eye feature map, a spatial Gaussian weighting function centered at the iris location is applied. Let $(c_x, c_y)$ denote the projected iris center coordinates in feature map space. The 2D Gaussian weight at spatial location $(x,y)$ is defined as:

\begin{equation}
G(x,y) = 
\frac{
\exp\left(
-\frac{(x - c_x)^2 + (y - c_y)^2}{2\sigma^2}
\right)
}{
\sum_{i=1}^{H} \sum_{j=1}^{W}
\exp\left(
-\frac{(i - c_x)^2 + (j - c_y)^2}{2\sigma^2}
\right)
}
\end{equation}

where, $\sigma$ (= 1.2 in our experiment) controls the spread of the Gaussian distribution. This normalized weighting map assigns higher importance to features closer to the iris center while suppressing peripheral regions.

The Gaussian-weighted feature map is then computed as:

\begin{equation}
\tilde{F}_e(c,x,y) = \hat{F}_e(c,x,y) \cdot G(x,y)
\end{equation}

where, $c$ denotes the channel index. Finally, a global representation is obtained via adaptive global average pooling: 

\begin{equation}
    E = \mathrm{GAP}(\tilde{F}_e) \in \mathbb{R}^{256}
    \label{eq:embedding_gap}
\end{equation}
where, $\mathrm{GAP}(\tilde{F}_e)$ is computed using similar Equation \ref{eq:gap_feature}. Note that $E$ is computed separately for the left and right eyes and denoted by $e_L$ and $e_R$. Figure \ref{fig:Eye Features of L4-Layer} provides the samples images for each step of eye feature extraction. This Gaussian-weighted global embedding enhances iris-centered discriminative information while retaining contextual eye features, making it particularly suitable for precise gaze estimation.

\subsubsection{Scene Feature Extraction}
The driver scene contains the contextual information of the driving environment, which influences the driver gaze. To extract this contextual information, we employ a pretrained ResNet-18 backbone as a spatial feature extractor, and features are extracted from the third residual block (Layer-3). Given a scene image of size $1280 \times 720$, the image is first resized \(
I_s \in \mathbb{R}^{3 \times 224 \times 224},
\) and normalized using ImageNet statistics before being forwarded through the convolutional stem and residual stages of ResNet-18. Unlike standard classification settings, the global average pooling and fully connected layers are removed in order to preserve spatial structure. The backbone therefore, produces a deep spatial feature map:

\begin{equation}
F_s = \mathcal{B}_3(I_s), 
\quad F_s \in \mathbb{R}^{256 \times 14 \times 14}
 \label{eq:scene_feature_map}
\end{equation}

where, $\mathcal{B}_3(\cdot)$ represents the convolutional feature extractor and the spatial resolution is reduced to $14 \times 14$ due to progressive down sampling. 


The spatial feature map is then reshaped into a sequence of grid-based tokens by flattening the spatial dimensions, $T_s = \text{reshape}(F_s) \in \mathbb{R}^{196 \times 256}$, where each of the 196 tokens corresponds to a spatial scene grid in the $14 \times 14$ feature map. More formally, for spatial location $(i,j)$,

\begin{equation}
t_{k} = F_s(:, i, j), 
\quad k = 14(i-1) + j
\end{equation}

with $t_k \in \mathbb{R}^{256}$ representing the contextual embedding of a specific scene region. These grid-based scene tokens preserve spatial correspondence between image regions and feature embeddings, enabling the subsequent Multi-Modal Feature Fusion module to regions about, where the driver is likely looking within the scene. The resulting representation $T_s \in \mathbb{R}^{196 \times 256}$ is stored and used as structured spatial context for gaze prediction.

\subsection{Multi-Modal Feature Fusion}
The multi-modal feature fusion refers to the process of combining features from multiple input modalities such as face appearance, eye features, iris position, and scene context into a unified representation that captures the gaze information. This fused representation enables the model to support complementary cues from different sources to improve the accuracy of gaze estimation
\subsubsection{Facial Features Representation}
\textit{Face Feature Encoding:} Face features extracted using four different layers of a CNN backbone are computed using Equation \ref{eq:gap_feature} and are represented as four global feature vectors
\(f_1, f_2, f_3, f_4 \in \mathbb{R}^{256}\).
 These features are concatenated and projected into a unified face embedding using the following equation: 

\begin{equation}
\mathbf{z}_{face} =
\mathrm{LN}\left(
\mathrm{ReLU}
\left(
W_{face}
\begin{bmatrix}
f_1 \\
f_2 \\
f_3 \\
f_4
\end{bmatrix}
+ b_{face}
\right)
\right)
\in \mathbb{R}^{256}
\label{Face_Features_Concatenation}
\end{equation}
where, $\mathrm{LN}(\cdot)$ denotes Layer Normalization, ReLU is the activation function, and \(W_{face}\) and \(b_{face} \) are the learnable weight matrix and bias vector respectively.\\

\textit{Eye Feature Fusion:} Left and right eye features \(
e_L, e_R \in \mathbb{R}^{256}
\), computed using Equation \ref{eq:embedding_gap} are concatenated and embedded as:
\(
\mathbf{z}_{eye} \in \mathbb{R}^{256}
\) using the similar as shown in Equation \ref{Face_Features_Concatenation}.
This representation captures fine-grained eye-driven gaze cues.\\

\textit{Iris Feature Embedding:} Iris centers from both eyes are represented as normalized coordinates:
\begin{equation}
\mathbf{\textit{$i_c$}} = [c_{x,L}, c_{y,L}, c_{x,R}, c_{y,R}] \in \mathbb{R}^4
\end{equation}
where, $(c_{x,L}, c_{y,L})$ and $(c_{x,R}, c_{y,R})$ represent the left and right iris center coordinates respectively.

These are projected to capture subtle eye movement corrections:
\begin{equation}
\mathbf{z}_{iris} =
\mathrm{LN}\left(
\mathrm{ReLU}
\left(
W_{iris}\mathbf{\textit{$i_c$}} + b_{iris}
\right)
\right)
\in \mathbb{R}^{256}
\end{equation}

\subsubsection{Gaze Intent Representation}

The modality-specific embeddings are concatenated to form a joint gaze descriptor:
\begin{equation}
\mathbf{z}_{cat} =
\begin{bmatrix}
\mathbf{z}_{face} \\
\mathbf{z}_{eye} \\
\mathbf{z}_{iris}
\end{bmatrix}
\in \mathbb{R}^{768}
\end{equation}

A gaze encoder compresses this representation into a unified gaze-intent vector:
\begin{equation}
\mathbf{z}_{gaze} =
\mathrm{LN}\left(
\mathrm{ReLU}
\left(
W_{gaze}\mathbf{z}_{cat} + b_{gaze}
\right)
\right)
\in \mathbb{R}^{256}
\end{equation}

This vector represents the driver latent gaze intention and serves as the query in the attention computation over the scene grid.

\subsubsection{Scene Representation}
The scene image is passed through a pretrained ResNet-18 backbone, producing a spatial feature map of size $14 \times 14$ as per Equation \ref{eq:scene_feature_map}. This results in 196 spatial features 
$\mathbf{s}_i \in \mathbb{R}^{256}, \; i = 1,\dots,196$, 
each corresponding to a different grid of the scene. To incorporate spatial information, each feature is associated with a normalized grid center 
$\mathbf{c}_i = (x_i, y_i)$, where $x_i, y_i$ are normalized to lie between 0 and 1. The horizontal position is encoded by concatenating the $x$-coordinate with the feature vector:
\begin{equation}
\tilde{\mathbf{s}}_i = [\mathbf{s}_i \;||\; x_i]
\end{equation}

\subsection{Gaze Prediction Head}

\subsubsection{3D Gaze Direction Estimation}
The 3D gaze direction is regressed directly from the fused gaze embedding 
$\mathbf{z}_{gaze} \in \mathbb{R}^{256}$. 
A fully connected layer projects this embedding into a 3-dimensional vector:

\begin{equation}
\hat{\mathbf{g}} = W_d \mathbf{z}_{gaze} + b_d
\end{equation}

where, $W_d \in \mathbb{R}^{3 \times 256}$ and $b_d \in \mathbb{R}^{3}$ are learnable parameters. Since gaze direction represents orientation in 3D space, the predicted vector is normalized to unit length:

\begin{equation}
\hat{\mathbf{g}} =
\frac{\hat{\mathbf{g}}}{\|\hat{\mathbf{g}}\|_2}
\end{equation}

Similarly, the ground-truth gaze vector $\mathbf{g}$ is normalized:

\begin{equation}
\mathbf{g} =
\frac{\mathbf{g}}{\|\mathbf{g}\|_2}
\end{equation}

\subsubsection{Attention based Point of Gaze Estimation}
PoG estimation involves fusing the face and scene grid, therfore to achieve this gaze intent vector, $\mathbf{z}_{gaze}$ is obtained from face image and used as a query vector, while the scene features are projected into key vectors $\mathbf{k}_i = W_k \tilde{\mathbf{s}}_i$, where, $W_k$ is a learnable projection matrix and $\tilde{\mathbf{s}}_i$ represents the feature vector of the $i$-th scene grid. Each key, $\mathbf{k}_i$ therefore corresponds to a spatial region in the scene represented by a grid with a predefined center coordinate. Attention scores between query vector and key vector are computed via dot-product similarity:

\begin{equation}
\mathbf{\alpha_i} = \frac{\exp(\mathbf{z}_{gaze}^\top \mathbf{k}_i)}
{\sum_{j=1}^{N} \exp(\mathbf{z}_{gaze}^\top \mathbf{k}_j)}
\end{equation}

The attention weight $\mathbf{\alpha_i}$ reflects the relevance between the gaze representation and the $i$-th scene grid. The final point-of-gaze is then obtained as the weighted average of the predefined grid center coordinates of the scene regions.
\begin{equation}
    \mathbf{\hat{{p}}} =
\sum_{i=1}^{N} \mathbf{\alpha_i} \mathbf{c}_i
\end{equation}

where $\mathbf{c}_i = (x_i, y_i)$ represents the normalized center coordinates of the $i^{\text{th}}$ scene grid in the scene image, and $i = 1, 2, \dots, 49$. \\

\textit{Vertical Residual Correction:} To compensate for grid discretization and vertical bias, a learnable residual is applied:
\begin{equation}
    \Delta \mathbf{p} = \lambda \tanh(W_p \mathbf{z}_{gaze})
\end{equation}

where $\lambda$ is a learnable scaling factor.

The final gaze prediction is:
\begin{equation}
\mathbf{\hat{P}}_{final} = \text{clip}(\mathbf{\hat{p}} + \Delta \mathbf{p}, 0, 1)
\end{equation}

\subsubsection{Loss Functions}
The loss of 3D gaze direction and PoG is computed using following loss function.\\

\textbf{\textit{Gaze Direction Loss:}} To ensure stable convergence and accurate angular prediction, we adopt a hybrid cosine and Euclidean loss.

First, cosine similarity between predicted and ground-truth gaze vectors is computed as:

\begin{equation}
\mathcal{L}_{cos} = 1 - (\hat{\mathbf{g}} \cdot \mathbf{g})
\end{equation}

This directly penalizes angular deviation. Additionally, an $L_2$ regression term is introduced:

\begin{equation}
\mathcal{L}_{L2} = \|\hat{\mathbf{g}} - \mathbf{g}\|_2
\end{equation}

The final direction loss is defined as:

\begin{equation}
\mathcal{L}_{dir} =
\lambda_1 \, \mathcal{L}_{cos}
+
\lambda_2 \, \mathcal{L}_{L2}
\end{equation}

where $\lambda_1 = 0.7 $ and $\lambda_2 = 0.3$ are weighting coefficients that balance the contribution of the cosine similarity loss $\mathcal{L}_{cos}$ and the Euclidean (L2) loss $\mathcal{L}_{L2}$ in the gaze direction supervision. This hybrid formulation improves training stability while preserving strong angular supervision.\\

\textbf{\textit{Point-of-Gaze Regression Loss:}}
We have used a Smooth L1 loss for PoG estimation. Smooth L1 loss is a combination of L1 loss (absolute error) and L2 loss (squared error). It behaves like L2 loss for small errors and like L1 loss for large errors. In our case, where the dataset is dominated by forward-looking samples, it helps reduce the impact of large errors from less-represented regions while maintaining accuracy for small errors. Let $\hat{\mathbf{p}} = (\hat{u}, \hat{v})$ denote the predicted normalized PoG and 
$\mathbf{p} = (u, v)$ the ground-truth normalized PoG.
We supervise PoG estimation using Smooth $L_1$ loss:

\begin{equation}
    \mathcal{L}_{pog} =
\begin{cases}
\frac{1}{2\beta} \|\hat{\mathbf{p}} - \mathbf{p}\|_2^2, & \|\hat{\mathbf{p}} - \mathbf{p}\| < \beta \\
\|\hat{\mathbf{p}} - \mathbf{p}\|_1 - \frac{\beta}{2}, & \text{otherwise}
\end{cases}
\end{equation}

A small transition parameter $\beta = 0.02$ is used to emphasize fine-grained spatial refinement.

\subsubsection{Evaluation Metrics}

\textbf{\textbf{Angular Error:}} Performance of 3D gaze direction estimation is evaluated using mean angular error (in degrees). 
The angular deviation between predicted and ground-truth vectors is computed as:

\begin{equation}
\theta =
\cos^{-1}
\left(
\text{clip}(\hat{\mathbf{g}} \cdot \mathbf{g}, -1, 1)
\right)
\times
\frac{180}{\pi}
\end{equation}

The final reported metric is the average angular error of all set of test data of all samples.\\


\textit{\textbf{Pixel Error:}} Predicted normalized PoG coordinates are converted to pixel space: $u_{px} = \hat{u} \cdot W_{img} ,
 \quad
v_{px} = \hat{v} \cdot H_{img}$ where, $W_{img}$ and $H_{img}$ denote image width and height, respectively.

Pixel error is then computed as:

\begin{equation}
\mathcal{E}_{pixel}
=
\|
\hat{\mathbf{p}}_{px}
-
\mathbf{p}_{px}
\|_2
\end{equation}

The reported value corresponds to the mean Euclidean distance in pixels across all set of test data of all samples.

\subsection{Implementation Details}
The proposed model is trained using the PyTorch framework on a GPU server with four NVIDIA GeForce RTX 3080 GPUs (10 GB VRAM each) and CUDA 11.4. Training is performed using the AdamW optimizer with a learning rate of $1 \times 10^{-4}$ and a weight decay of $1 \times 10^{-4}$. The model is trained for 40 epochs with a batch size of 64. To improve training stability, gradient clipping with a maximum norm of 5.0 is applied during backpropagation. The model also employs dropout with a probability of 0.3 within the gaze encoding module to reduce overfitting. Early stopping is used to prevent overfitting, with a patience of 6 epochs based on validation error. The model parameters corresponding to the lowest validation pixel error are saved as the final model checkpoint and used for evaluation.


\section{Results and Discussion}
In this section, we first discuss the performance of the 3D gaze direction estimation model and compare its results with state-of-the-art (SOTA) models. Next, we evaluate the SGAP-Gaze model for Point-of-Gaze (PoG) estimation and compare it's performance with a benchmark gaze estimation model, GazePTR, followed by a discussion on the failure cases and limitations of the proposed models.

\subsection{3D Gaze Direction Estimation}
3D gaze direction estimation aims to predict the orientation of the driver's gaze vector in three-dimensional space. The gaze vector indicates the direction the driver is looking and is often used to generate scene saliency maps that indicate potential gaze regions in the scene image. However, a scene saliency map typically provides approximate gaze areas in the scene image. To estimate 3D gaze direction in this study, we used the benchmark LBW dataset, which contains 3D gaze ground-truth vectors. A subject-independent evaluation protocol is adopted to evaluate the generalization capability of our proposed model. The dataset is split at the driver level to ensure that training, validation, and test samples come from different individuals. The data from the first 17 drivers are used for training, the next 5 for validation, and the remaining 6 for testing. This setup ensures that the model is evaluated on previously unseen drivers, providing a realistic assessment of its performance in real-world scenarios. 

\par Table \ref{tab:comparison_results} shows the performance of our proposed 3D gaze direction estimation model along with other SOTA models, tested on the benchmark LBW dataset. The dataset setup is similar to that used by \citet{kasahara2022look}, which helps in fair comparison with existing state-of-the-art methods. This includes testing our proposed model with only the driver's face image as input, without scene information, as shown in Table \ref{tab:comparison_results}. Early methods such as Gaze360 and ETH-XGaze report mean angular errors of 20.3° and 15.6°, respectively. More recent transformer-based approaches have significantly improved performance, with GazePTR achieving 6.18° mean angular error. The LBW method reports a slightly higher error of 6.70°. Our proposed method achieves a mean angular error of 6.04°, thereby improving on the existing benchmark model performance.

\begin{table*}[!ht]
\centering
\caption{Results of 3D gaze direction estimation with SOTA models}
\label{tab:comparison_results}
\begin{tabular}{c c c}
\hline
Method      & Input & Mean   Angular Error \\ \hline
Gaze360     & F     & 20.3                 \\
ETH   Xgaze & F     & 15.6                 \\
GazePTR     & F     & 6.18                 \\
LBW         & F     & 6.70                 \\ \hline
Proposed         & F     & 6.04                 \\ \hline
\end{tabular}
\end{table*}

\subsection{Point of Gaze Estimation}
Although 3D gaze direction estimation provides information about the driver's gaze orientation in space, it only indicates the direction of gaze. It does not identify the exact location in the scene being observed. Point-of-Gaze (PoG) estimation addresses this limitation by projecting the gaze onto the scene image to determine the precise spatial location where the driver is looking. This enables a more accurate understanding of driver attention within the driving environment. Our proposed point-of-gaze model SGAP-Gaze is evaluated primarily on our benchmark UD-FSG dataset. This is because our UD-FSG dataset contains more complex and challenging traffic scenarios than the LBW dataset and is better suited for developing a scene-grid attention-based PoG prediction model. However, to further assess the generalization ability of our PoG model, we also evaluate its performance on the benchmark LBW dataset for comparison.  

\subsubsection{Overall Performance}
In this study, the proposed Point-of-Gaze (PoG) estimation model SGAP-Gaze is evaluated using our UD-FSG dataset. The driver-independent evaluation protocol is followed for training, validation, and testing the proposed model, which is similar to the training protocol used for 3D gaze direction estimation, as discussed earlier. The model achieved an overall mean pixel error of 96.16 on the test data, which is approximately 6.55\% of the diagonal size of the scene image ($1280 \times 720$ pixels). Please note that all results from Point-of-Gaze estimation experiments on the UD-FSG dataset are based on input from facial features (including face, eye, and iris) and scene features.

\par To further analyse the performance of our proposed model, several experiments have been conducted to determine the contribution of different input modalities: face features (F), face and eye features (F+E), and face, eye, and scene features (F+E+S). The results are summarized in Table \ref{tab:comparison_results_input_features}. It can be seen that a mean pixel error of 108.63 pixels with a standard deviation (SD) of 92.58 pixels is obtained using only facial features. When iris-weighted eye features are incorporated with face features, the mean error decreases to 105.47 pixels, and the standard deviation further decreases to 84.87, suggesting that eye information provides more reliable gaze cues and improves prediction consistency. Further integrating scene features reduces the mean pixel error to 96.16 pixels and the standard deviation to 75.85, demonstrating that scene context helps the model better localize the point-of-gaze while also improving the stability of the predictions, as reflected in the reduction in SD values. 

\par The above analysis shows that the scene grid attention mechanism helps distribute predicted PoG across different spatial regions of the scene images rather than concentrating only near the center. The progressive reduction in both mean error and standard deviation highlights the effectiveness of combining face, eye, and scene information for accurate and consistent PoG estimation. The median pixel error of 76.92 pixels further supports this observation, as it represents the typical performance of the model and is notably lower than the mean error. The noticeable gap between the median and mean values suggests that the distribution is influenced by a limited number of high-error outliers, which shift the mean toward larger values and contribute to the high standard deviation.

\begin{table*}[ht]
\centering
\caption{Comparison of PoG estimation on UD-FSG dataset using different input modalities}
\label{tab:comparison_results_input_features}
\begin{tabular}{c c c}
\hline
Input & Mean   Pixel Error & SD   \\ \hline
Face     & 108.63             & 92.58 \\
Face+Eye   & 105.47             & 84.87 \\
Face+Eye+Scene & 96.16             & 75.85 \\
\hline
\end{tabular}
\end{table*}

\par To check the generalization capability of the proposed SGAP-Gaze model across different traffic environments, it is further evaluated by training and testing on the LBW dataset. The overall performance of the proposed model on the LBW and UD-FSG datasets is summarized in Table \ref{tab:pixel_error_on_LBW_UD_FSG}. On the LBW dataset, the model achieves a mean pixel error of 63.48 pixels, corresponding to a normalized error of 5.98\%. Similarly on the UD-FSG dataset, the model obtains a mean pixel error of 96.16 pixels with a normalized error of 6.55\%. Here, the normalized error is computed by dividing the mean pixel error by the diagonal (in pixels) of the scene image as shown in Table \ref{tab:pixel_error_on_LBW_UD_FSG}. The normalized error results show that the model performance is slightly better on LBW dataset compared with UD-FSG. However, the higher pixel error observed on the UD-FSG dataset can be attributed to its higher image resolution and higher scene complexity, which introduce greater spatial variability in gaze targets. Despite these challenges, the overall normalized error remains relatively low across both datasets demonstrating the robustness of the proposed SGAP-Gaze in estimating the driver’s Point-of-Gaze across different datasets and scene conditions.

\begin{table*}[ht]
\centering
\caption{Overall performance of SGAP-Gaze on LBW and UD-FSG datasets}
\label{tab:pixel_error_on_LBW_UD_FSG}
\begin{tabular}{c c c c c}
\hline
Dataset & Resolution & Diagonal & Mean   Pixel Error & Normalized   Error (\%) \\ \hline
LBW     & 942 × 489    & 1061     & 63.48              & 5.98                    \\
UD-FSG  & 1280 × 720   & 1468     & 96.16            & 6.55                    \\ \hline
\end{tabular}
\end{table*}

\subsubsection{Comparative Analysis of SGAP-Gaze with GazePTR Model}

For consistency, GazePTR is selected as the baseline for the PoG comparison, while GazePTR, GLGaze, and LNet are considered in the 3D gaze-direction comparison. The two evaluations are treated separately because PoG localization and 3D gaze-direction estimation represent different prediction tasks and use different evaluation metrics.
Therefore, we perform further analysis by considering GazePTR as the baseline model for \textbf{point-of-gaze (PoG) estimation} and compare its performance with our proposed SGAP-Gaze. The corresponding results are presented in Table \ref{tab:comparision_of_SGAP-Gaze_with_Gaze-PTR}. GazePTR achieves a mean pixel error of 136.96 pixels with a standard deviation (SD) of 95.44 over 12,571 samples. In contrast, the proposed SGAP-Gaze achieves mean pixel error of 96.16 pixels, with a standard deviation of 75.85, on the same dataset. This corresponds to a reduction of 40.8 pixels in mean pixel error, representing an improvement of approximately 29.78\% compared to GazePTR. Additionally, the lower SD indicates that SGAP-Gaze produces more stable, consistent gaze predictions than GazePTR. This improvement can be attributed to the proposed scene grid attention mechanism, which effectively integrates facial features (from face, eye, iris) with scene contextual information to better localize the driver's point of gaze in the scene.

\begin{table*}[ht]
\centering
\caption{Performance comparison of SGAP-Gaze with GazePTR in PoG estimation}
\label{tab:comparision_of_SGAP-Gaze_with_Gaze-PTR}
\begin{tabular}{c c c c}
\hline
Method  & Mean   Pixel Error & SD    & Sample   Size \\ \hline
GazePTR & 136.96             & 95.44 & 12571         \\
SGAP-Gaze (Our)     & 96.16             & 75.85 & 12571         \\ \hline
\end{tabular}
\end{table*}

\par The cumulative accuracy results in Table \ref{tab:Cumulative_Pixel_Error} demonstrate the superior performance of SGAP-Gaze compared to GazePTR across all pixel error thresholds. SGAP-Gaze achieves significantly higher accuracy at lower error levels, with 29.44\% of predictions within 50 pixels, compared to 14.34\% for GazePTR. Similarly, 63.61\% of SGAP-Gaze predictions fall within 100 pixels, whereas GazePTR achieves only 43.56\%, highlighting a substantial improvement in PoG estimation. The consistent advantage of SGAP-Gaze across all thresholds indicates that the proposed scene grid attention mechanism enables more accurate and reliable Point-of-Gaze estimation. This improvement is particularly important for driver attention analysis, where accurate localization of gaze targets is required.

\begin{table*}[ht]
\centering
\caption{Cumulative accuracy of PoG estimation at different pixel error levels}
\label{tab:Cumulative_Pixel_Error}
\begin{tabular}{ccccc}
\hline
\multirow{2}{*}{Pixel   Error} & \multicolumn{2}{c}{GazePTR} & \multicolumn{2}{c}{SGAP-Gaze} \\
                               & Count   & Accuracy (\%)  & Count   & Accuracy   (\%)   \\ \hline
\textless{}50                  & 1803    & 14.34             & 3701    & 29.44              \\
\textless{}96                 & 5144    & 40.92             & 7710    & 61.33              \\
\textless{}100                 & 5476    & 43.56             & 7997    & 63.61              \\
\textless{}125                 & 6802    & 54.11             & 9401    & 74.78              \\
\textless{}150                 & 8219    & 65.38             & 10372    & 82.50              \\
\textless{}200                 & 10183   & 81.00             & 11518   & 91.62              \\
\textless{}500                 & 12470   & 99.19             & 12531   & 99.68              \\ \hline
\end{tabular}
\end{table*}

\subsubsection{Spatial Pixel Error Distribution Analysis}
\begin{figure}[!htbp]
    \centering
    \includegraphics[width=0.65\textwidth]{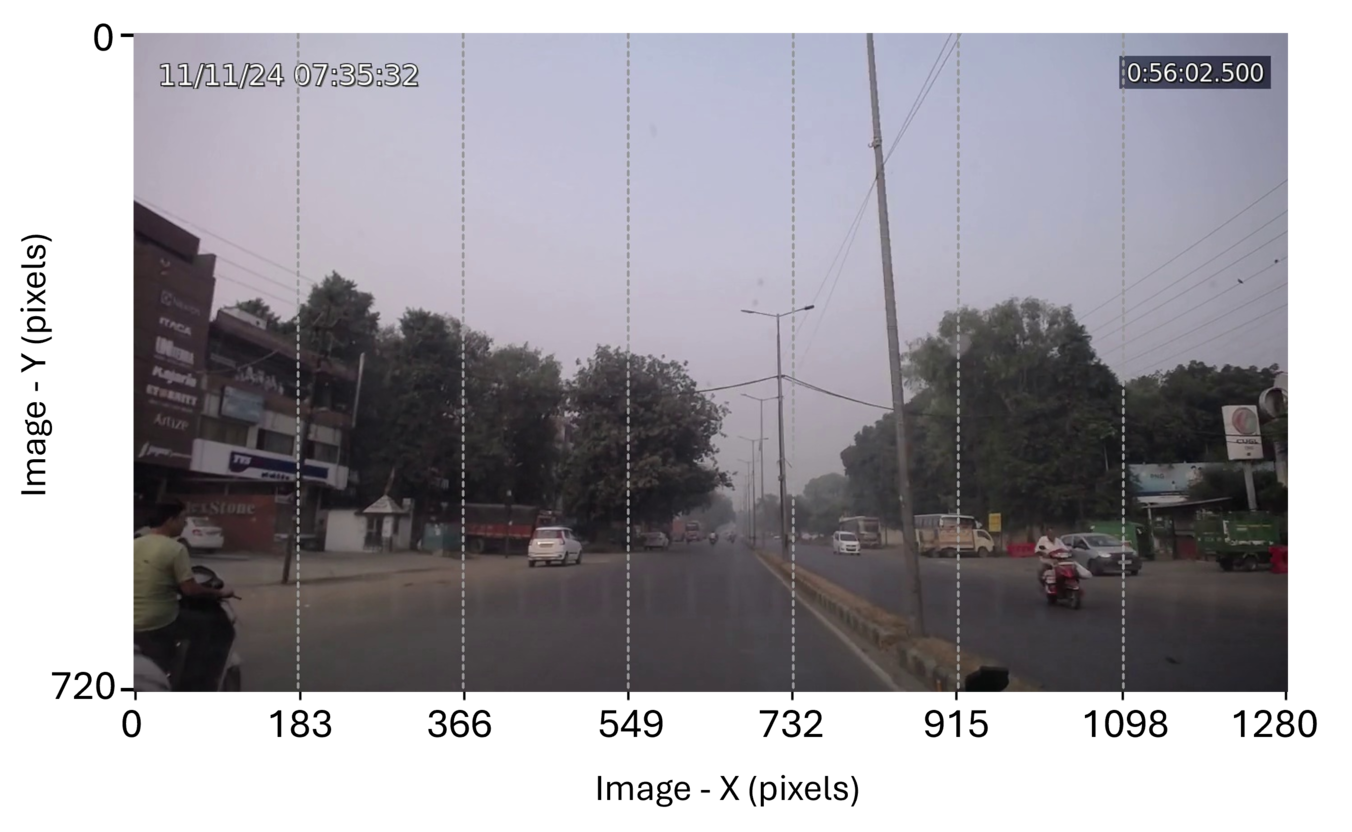}
     \caption{Sample image showing division of  scene image width into seven equal parts.}
    \label{fig:divide_image}
\end{figure}

The performance of SGAP-Gaze and GazePTR is further analyzed across different spatial location in the scene image. First, the scene image (1280 × 720) is horizontally divided into seven equal regions based on the x-coordinate of the gaze ground-truth (Gaze GT-X) to analyze the spatial behavior of the models. Figure \ref{fig:divide_image} shows an example of scene image horizontally divided into seven equal parts. The mean pixel error is computed for each region for both GazePTR and SGAP-Gaze, as presented in Table \ref{tab:spatial_pixel_error_distribution}. The results show that SGAP-Gaze consistently achieves lower mean pixel error across all spatial regions compared to GazePTR. In the outer regions of the scene (0–183 and 1098–1280), where gaze targets are located near the scene boundaries, SGAP-Gaze shows lower mean pixel errors (327.70 and 352.72 respectively) compared to GazePTR (447.7 and 444.29). However, performance in these outer regions still requires improvement, as the errors remain significantly higher than those observed in the central regions. 
 In the central regions of the image (366–915), where most gaze samples are concentrated, SGAP-Gaze still achieves consistently lower errors than GazePTR, demonstrating improved accuracy and stability in gaze localization. The lower error in the central region for both models is mainly due to the higher density of gaze samples, as drivers predominantly look forward. The fact that the driver gaze is predominantly in the forward regions is also supported by the study by \citet{fu2012automatic}. Overall, these results highlight the effectiveness of the grid-based attention mechanism in improving spatial robustness for Point-of-Gaze estimation as compared to existing state-of-the-art methods.

\begin{table*}[ht]
\centering
\caption{Comparison of spatial pixel error distribution of GazePTR and SGAP-Gaze}
\label{tab:spatial_pixel_error_distribution}
\begin{tabular}{ccccccc}
\hline
\multirow{2}{*}{Range (Gaze GT-X)} & \multicolumn{2}{c}{GazePTR} & \multicolumn{2}{c}{SGAP-Gaze} & \multirow{2}{*}{Count} & \multirow{2}{*}{Sample (\%)} \\
                                   & Mean Pixel Error  & SD      & Mean Pixel Error  & SD       &                        &                              \\ \hline
0-183                              & 447.7             & 137.55  & 327.70            & 126.16   & 60                     & 0.47                         \\
183-366                            & 261.27            & 110.1   & 242.61           & 186.49  & 203                    & 1.61                         \\
366-549                            & 155.38            & 71.64   & 96.72            & 64.53    & 908                    & 7.22                         \\
549-732                            & 104.11            & 68.12   & 83.01             & 58.11    & 8458                   & 67.28                        \\
732-915                            & 181.39            & 88.42   & 103.99            & 68.01    & 2574                   & 20.47                        \\
915-1098                           & 251.27            & 130.99  & 202.88            & 100.63   & 317                    & 2.52                         \\
1098-1280                          & 444.29            & 168.9   & 352.72           & 122.64  & 51                     & 0.41                         \\ \hline
\end{tabular}
\end{table*}

\subsubsection{Visualization of Predicted and Ground Truth Point-of-Gaze (PoG)}
\begin{figure}[!htbp]
    \centering
    \includegraphics[width=1.0\textwidth] {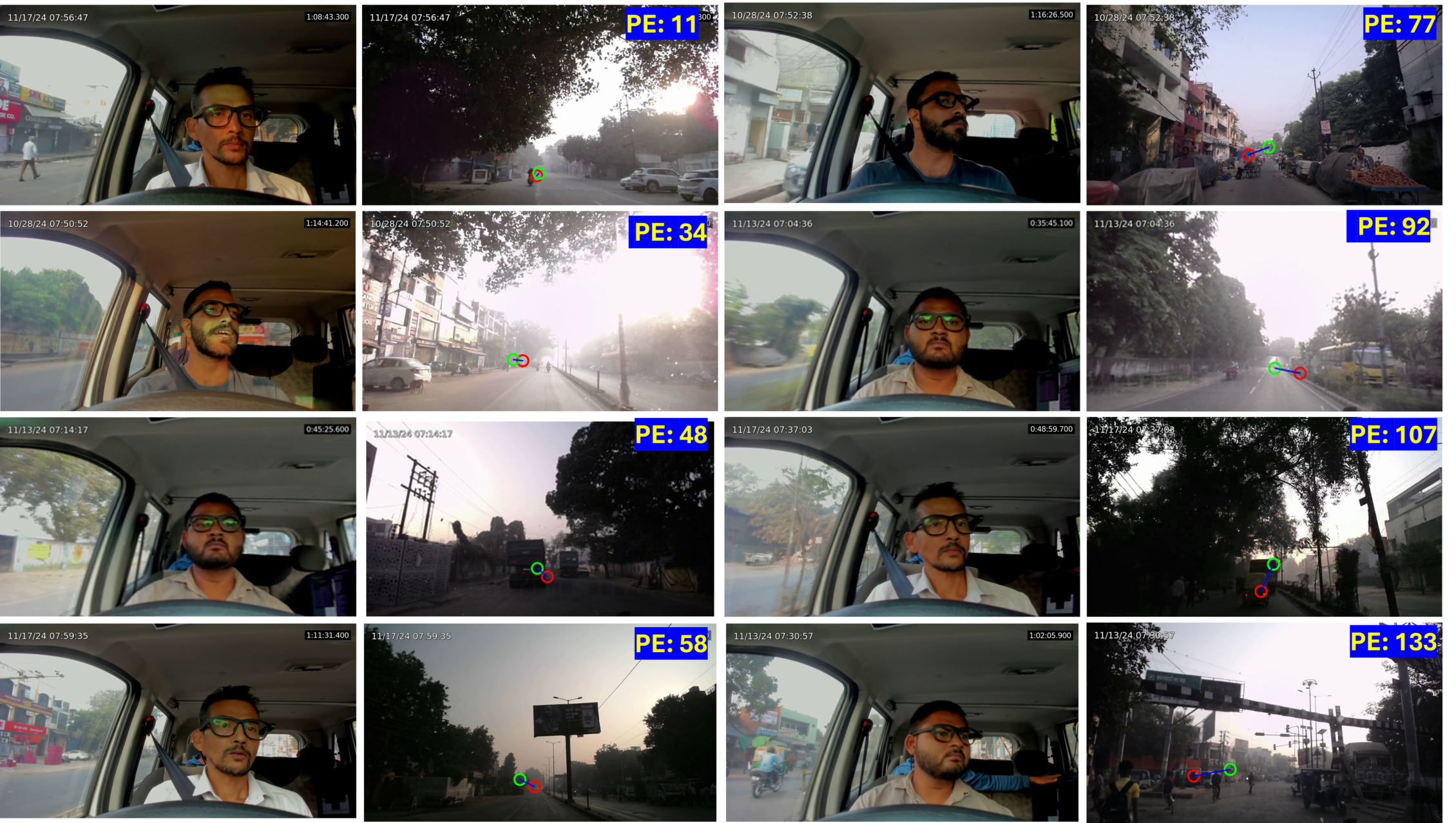}
     \caption{PoG prediction results on test samples. The green circle and red circle (radius = 20 pixels) represent the ground-truth and predicted PoG, respectively.}
    \label{fig:Predicted_Ground_Truth_Result}
\end{figure}

The visualization of the predicted PoG and ground-truth PoG along with the corresponding driver face helps to understand the qualitative performance of the model. Figure \ref{fig:Predicted_Ground_Truth_Result} shows a few samples of PoG predictions with varying error magnitudes. The green circle denotes the ground-truth PoG, while the red circle denotes the PoG predicted using our proposed SGAP-Gaze model. For ease of visualization, we have used a circle of radius 20 pixels around the PoG.

\par In the top-left sample, the model achieves high accuracy with a pixel error of 11 pixels, indicating that the predicted gaze location is very close to the ground truth. In contrast, the bottom-right sample shows a larger error of 133 pixels, where the predicted gaze location deviates significantly from the actual gaze point. These higher errors typically occur when the driver looks toward the edges of the scene. In comparison, lower errors are commonly observed in the central region of the scene, where drivers predominantly focus while driving.
This observation is also supported by the quantitative results in Table \ref{tab:spatial_pixel_error_distribution}. 
These results indicate that the proposed model achieves reliable gaze localization in most cases, while larger errors mainly occur for gaze targets located near the scene boundaries.

\par \textit{Failure Cases:} The mean pixel error of SGAP-Gaze evaluated on 12,571 face–scene image pairs from unseen drivers is 96.16 pixels. As shown in Table \ref{tab:Cumulative_Pixel_Error}, approximately 61.33\% of the predictions have errors below 96 pixels, and about 90\% of the test samples have errors below 200 pixels, while only around 10\% of the cases exhibit errors greater than 200 pixels.
Several factors contribute to these failure cases where the predicted PoG deviates significantly from the ground truth. One common reason for errors greater than 200 pixels is the difficulty in accurately detecting the driver’s iris, particularly when the face image quality is degraded due to motion blur, low illumination, or reflections from sunlight on the eye-tracker glasses. In such situations, the model relies primarily on head orientation cues, while the eye-based gaze information becomes unreliable or unavailable. Another failure scenario occurs during eye blinks, where the iris is temporarily occluded, causing the model to estimate gaze mainly from facial orientation rather than precise eye cues. This limitation could potentially be addressed by incorporating temporal information across consecutive frames, which is not considered in the current model.

\section{Conclusion}

This study developed a Point-of-Gaze (PoG)-based driver gaze estimation framework. To support this framework, we used our a large-scale, real-world Urban Driving Face-Scene Gaze (UD-FSG) dataset collected in a heterogeneous traffic environment. The dataset comprises of synchronized driver face images, scene images, and corresponding 2D gaze ground truth, enabling comprehensive evaluation under realistic driving scenarios. 

\par We proposed a scene-grid attention-based point-of-gaze (SGAP-Gaze) network that computes attention between the gaze intent vector and scene information using a Transformer-based mechanism. Since the driver’s face image in the UD-FSG dataset includes surrounding context, a YOLOv8-based Face-Eye-Iris (FEI) detector is trained on approximately 481 drivers’ 2200 annotated face images. The FEI detector is used to detect driver face, eye, and iris from the driver face image captured using the dashboard webcam face camera. The detected face, eyes, and iris are then processed using ResNet-18 model to extract multi-layer facial features, along with iris-weighted eye features of Layer-4 to capture fine-grained eye movements. These features are then fused to form a gaze intent vector, which serves as the query for attention computation. Meanwhile, scene features are extracted from Layer-3 of ResNet-18 model and represented as a 14×14 feature map, which can be defined as a spatial grid of size 14×14. Then, attention scores are computed between the gaze intent vector and each scene grid to estimate the point of gaze. Besides the PoG based estimation, we also adopted our proposed model for 3D gaze direction using the face information only and compared with existing state-of-the-art models.

\par Our analysis shows that the 3D gaze direction model outperforms all existing benchmark models. We also performed a detailed analysis of the SGAP-Gaze model in terms of PoG estimation.  Experimental results show that SGAP-Gaze achieves a mean pixel error (MPE) of 96.16 on the UD-FSG dataset and 63.48 MPE on the LBW dataset, which is 6.55\% and 5.98\% in the normalized diagonal of the scene image.  Compared to the existing state-of-the-art driver gaze estimation model (GazePTR), our proposed model achieves a 29.78\% reduction in mean pixel error on the UD-FSG dataset. Further, the spatial pixel distribution analysis shows that SGAP-Gaze consistently achieves lower mean pixel error than GazePTR across all spatial ranges. The improvement is particularly significant in outer regions and near scene boundaries, indicating better robustness in challenging peripheral scene regions.

Despite these promising results, certain limitations of our model can be addressed or expanded upon in the future. In future work, this framework can be expanded by incorporating nighttime driving data to evaluate performance under low-light conditions. Additionally, the scene representation can explicitly include regions corresponding to left- and right-wing mirrors, which would help to generalize the model further. Finally, integrating intrinsic and extrinsic camera parameters can enhance the model's scalability across different vehicles, leading to large-scale real-world application.

\section*{CRediT authorship contribution statement}

\textbf{Pavan Kumar Sharma:} Conceptualization, Methodology, Data curation, Investigation, Validation, Writing – original draft, Writing – review \& editing; \textbf{Pranamesh Chakraborty:} Methodology, Supervision, Resources, Funding acquisition, Writing – review \& editing 

\section*{Declaration of competing interest}
The authors declare that they have no known competing financial interests or personal relationships that could have appeared to influence the work reported in this paper.


\section*{Acknowledgment}
We sincerely acknowledge Abhay Shukla, an undergraduate student, for his valuable assistance in transforming the ground-truth gaze coordinates from eye-tracker reference images to dashboard scene-camera images using homography matrices for the UD-FSG dataset.

\section*{Data availability}
The dataset supporting the findings of this study is available upon request through the following link:
\url{https://github.com/pavans20/Urban-Driving-Face-Scene-Gaze-Dataset.git}.

\bibliography{bibliography}
\end{document}